\documentclass[11pt,letterpaper]{article}
\usepackage{naaclhlt2015}
\usepackage{times}
\usepackage{latexsym}
\usepackage{amsthm}
\usepackage{amssymb}
\usepackage{amsmath}
\usepackage{tabulary}
\usepackage{graphicx}
\usepackage{booktabs} 
\usepackage{array} 
\usepackage{paralist} 
\usepackage{verbatim} 
\usepackage{subfig} 

\usepackage{amsfonts}
\usepackage{microtype}
\usepackage[scaled=0.86]{helvet}
\newcommand{\captionfonts}{\small}
\makeatletter  
\long\def\@makecaption#1#2{%
  \vskip\abovecaptionskip
  \sbox\@tempboxa{{\captionfonts #1: #2}}%
  \ifdim \wd\@tempboxa >\hsize
    {\captionfonts #1: #2\par}
  \else
    \hbox to\hsize{\hfil\box\@tempboxa\hfil}%
  \fi
  \vskip\belowcaptionskip}
\makeatother   

\belowdisplayskip \abovedisplayskip

\usepackage[nodisplayskipstretch]{setspace}
\newcommand{\B}[1]{#1}
\newtheorem*{prb-rank}{Context-Sensitive Ranking}
\newtheorem*{prb-gen}{Context-Sensitive Generation}

\title{A Neural Network Approach to \\ Context-Sensitive Generation of Conversational Responses\thanks{~~This paper appeared in the proceedings of NAACL-HLT 2015 (submitted December 4, 2014, accepted February 20, 2015, and presented June 1, 2015).}}
\author{Alessandro Sordoni\hspace{1pt}$^{{{\bf 1}}\dag\ddag}$ \hspace{.5cm} Michel Galley\hspace{1pt}$^{{\bf 2}\ddag}$  \hspace{.5cm}  Michael Auli\hspace{1pt}$^{{\bf 3}\dag}$  \hspace{.5cm}  Chris Brockett\hspace{1pt}$^{\bf 2}$ \\[0.05cm] {\bf Yangfeng Ji\hspace{1pt}$^{{\bf 4}\dag}$  \hspace{.45cm}  Margaret Mitchell\hspace{1pt}$^{\bf 2}$  \hspace{.45cm}  Jian-Yun Nie\hspace{1pt}$^{{\bf 1}\dag}$ \hspace{.45cm} Jianfeng Gao\hspace{1pt}$^{\bf 2}$ \hspace{.45cm}  Bill Dolan\hspace{1pt}$^{\bf 2}$} 
\\[0.3cm]
{$^1$DIRO, Universit\'e de Montr\'eal, Montr\'eal, QC, Canada} \\
{$^2$Microsoft Research, Redmond, WA, USA} \\
{$^3$Facebook AI Research, Menlo Park, CA, USA}\\
{$^4$Georgia Institute of Technology, Atlanta, GA, USA}\\
}
\date{}
\begin{document}	
\maketitle

{\let\thefootnote\relax\footnotetext{$\dag$The entirety of this work was conducted while at Microsoft Research.}}
{\let\thefootnote\relax\footnotetext{$^\ddag$Corresponding authors: Alessandro Sordoni (sordonia@iro.umontreal.ca) and Michel Galley (mgalley@microsoft.com).}}

\begin{abstract}
We present a novel response generation system that can be trained end to end on large quantities of unstructured Twitter conversations.  A neural network architecture is used to address sparsity issues that arise when integrating contextual information into classic statistical models, allowing the system to take into account previous dialog utterances. Our dynamic-context generative models show consistent gains over both context-sensitive and non-context-sensitive Machine Translation and Information Retrieval baselines.
\end{abstract}

\section{Introduction}
Until recently, the goal of training open-domain conversational systems that emulate human conversation has seemed elusive. However, the vast quantities of conversational exchanges now available on social media websites such as Twitter and Reddit raise the prospect of building data-driven models that can begin to communicate conversationally. 
The work of~\newcite{Ritter2011}, for example, demonstrates that a response generation system can be constructed from Twitter conversations using statistical machine translation techniques, where a status post by a Twitter user is ``translated'' into a  plausible looking response.

\begin{figure}[tbp]
\centering
\includegraphics[scale=0.7]{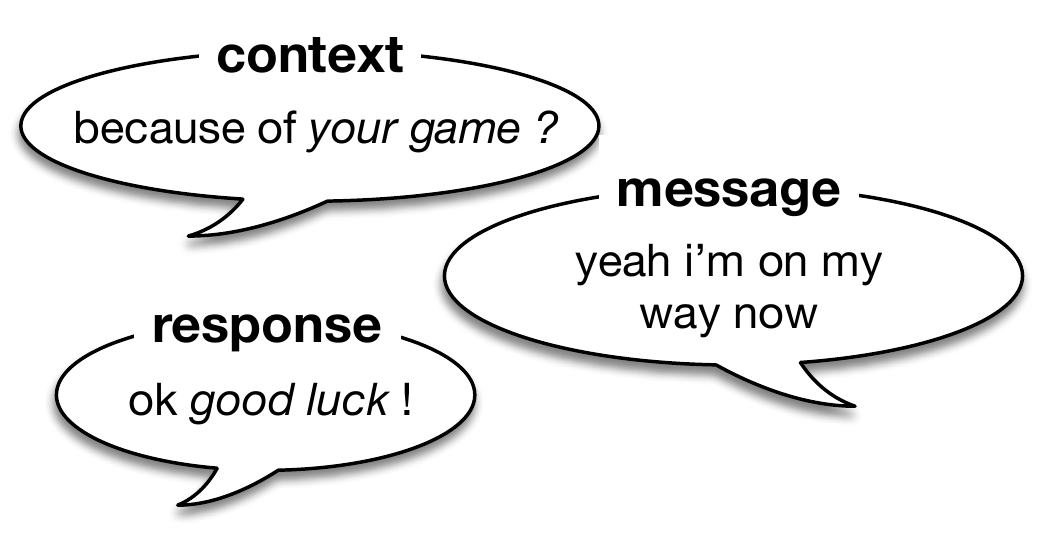}
\caption{\label{tab:sample_triple} 
Example of three consecutive utterances occurring between two Twitter users $A$ and $B$.}
\end{figure}

However, an approach such as that presented in~\newcite{Ritter2011} does not address the challenge of generating responses that are sensitive to the context of the conversation. Broadly speaking, context may be linguistic or involve grounding in the physical or virtual world, but we here focus on linguistic context. The ability to take into account previous utterances is  key to building dialog systems that can keep conversations active and engaging.  Figure~\ref{tab:sample_triple} illustrates a typical Twitter dialog where the contextual information is crucial: the phrase ``good luck'' is plainly motivated by the reference to ``your game'' in the first utterance.   In the MT model, such contextual sensitivity is difficult to capture;  moreover, naive injection of context information would entail unmanageable growth of the phrase table at the cost of increased sparsity, and skew towards rarely-seen context pairs.  In most statistical approaches to machine translation, phrase pairs do not share statistical weights regardless of their intrinsic semantic commonality.

We propose to address the challenge of context-sensitive response generation by using continuous representations or \emph{embeddings} of words and phrases to compactly encode semantic and syntactic similarity. We argue that embedding-based models afford flexibility to model the transitions between consecutive utterances and to capture long-span dependencies in a domain where traditional word and phrase alignment is difficult~\cite{Ritter2011}.  To this end, we present two simple, context-sensitive response-generation models utilizing the Recurrent Neural Network Language Model (RLM) architecture of~\cite{Mikolov2010}.  These  models first encode past information in a hidden continuous representation, which is then decoded by the RLM to promote plausible responses that are simultaneously fluent and contextually relevant. Unlike typical complex task-oriented multi-modular dialog systems~\cite{Young2002,StentBook2014}, our architecture is \emph{completely data-driven} and can easily be trained end-to-end using unstructured data without requiring human annotation, scripting, or automatic parsing.

This paper makes the following contributions. 
We present a neural network architecture for response generation that is both context-sensitive and data-driven. 
As such, it can be trained from end to end on massive amounts of social media data. 
To our knowledge, this is the first application of a neural-network model to open-domain response generation, and we believe that the present work will lay groundwork for more complex models to come.  
We additionally introduce a novel multi-reference extraction technique that shows promise for automated evaluation.

\section{Related Work}


Our work naturally lies in the path opened by~\newcite{Ritter2011}, but we generalize their approach by exploiting information from a larger context.
Ritter et al. and our work represent a radical paradigm shift from other work in dialog. 
More traditional dialog systems typically tease apart dialog management \cite{Young2002} from response generation \cite{StentBook2014}, while our holistic approach can be considered a first attempt to accomplish both tasks jointly.
While there are previous uses of machine learning for response generation \cite{Walker2003}, dialog state tracking \cite{Young2010}, and user modeling \cite{Georgila2006}, many components of typical dialog systems remain hand-coded: in particular, the labels and attributes defining dialog states. In contrast, the dialog state in our neural network model is completely latent and directly optimized towards end-to-end performance.
In this sense, we believe the framework of this paper is a significant milestone towards more data-driven and less hand-coded dialog processing.

Continuous representations of words and phrases estimated by neural network models 
have been applied on a variety of tasks ranging from Information Retrieval (IR)~\cite{Huang2013,Shen2014}, Online Recommendation~\cite{GaoOR14}, Machine Translation (MT)~\cite{Auli2013,Kyung14,Blunsom2013,Ilya2014}, and Language Modeling (LM)~\cite{Bengio2003,Collobert2008}.
Gao et al.~\shortcite{Gao2014} successfully use an embedding model to refine the estimation of rare phrase-translation probabilities, which is traditionally affected by sparsity problems. 
Robustness to sparsity is a crucial property of our method, as it allows us to capture context information while avoiding unmanageable growth of model parameters.

Our work extends the Recurrent Neural Network Language Model (RLM) of~\cite{Mikolov2010}, which uses continuous representations to estimate a probability function over natural language sentences. 
We propose a set of conditional RLMs where contextual information (i.e., past utterances) is encoded in a continuous context vector to help generate the response. 
Our models differ from most previous work in the way the context vector is constructed. For example, \newcite{Mikolov2012} and \newcite{Auli2013} use a pre-trained topic model.
In our models, the context vector is learned along with the conditional RLM that generates the response.
Additionally, the learned context encodings do not exclusively capture contentful words. Indeed, even ``stop words'' can carry discriminative power in this task; for example, all words in the utterance ``how are you?'' are commonly characterized as stop words, yet this is a contentful dialog utterance.

\section{Recurrent Language Model}
We give a brief overview of the Recurrent Language Model (RLM)~\cite{Mikolov2010} architecture that our models extend.
A RLM is a generative model of sentences, i.e., given sentence $s = s_1, \ldots, s_T$, it estimates: 
\begin{equation}
p(s) = \prod_{t = 1}^T p(s_t | s_{1}, \ldots, s_{t-1}).
\end{equation}
The model architecture is parameterized by three weight matrices, $\Theta_{\text{RNN}} = \langle \B W_{in}, \B W_{out}, \B W_{hh} \rangle$: an input matrix $ W_{in}$, a recurrent matrix $W_{hh}$ and an output matrix $\B W_{out}$, which are usually initialized randomly. 
The rows of the input matrix $\B W_{in} \in \mathbb{R}^{V \times K}$ contain the $K$-dimensional embeddings for each word in the language vocabulary of size $V$. Let us denote by $\B s_t$ both the vocabulary token and its one-hot representation, i.e., a zero vector of dimensionality $V$ with a $1$ corresponding to the index of the $s_t$ token. 
The embedding for $s_t$ is then obtained by $\B s_t^\top \B W_{in}$. 
The recurrent matrix $\B W_{hh} \in \mathbb{R}^{K \times K}$ keeps a history of the subsequence that has already been processed. 
The output matrix $\B W_{out} \in \mathbb{R}^{K \times V}$ projects the hidden state $h_t$ into the output layer $o_t$, which has an entry for each word in the vocabulary $V$. This value is used to generate a probability distribution for the next word in the sequence. Specifically, the forward pass proceeds with the following recurrence, for $t = 1, \ldots, T$:
\begin{equation}
\B h_t = \sigma( \B s_t^\top \B W_{in} + \B h_{t-1}^\top \B W_{hh} ),\;\; \B o_t = \B h_t^\top \B W_{out}
\end{equation}
where $\sigma$ is a non-linear function applied element-wise, in our case the logistic sigmoid. 
The recurrence is seeded by setting $\B h_0 = 0$, the zero vector.
The probability distribution over the next word given the previous history is obtained by applying the softmax activation function:
\begin{equation}
P(s_{t} = w | s_{1}, \ldots, s_{t-1}) = \frac{\exp (o_{tw})}{\sum_{v=1}^V \exp (o_{tv})}.
\end{equation}
The RLM is trained to minimize the negative log-likelihood of the training sentence $\B s$:
\begin{equation}
L(s) = - \sum_{t=1}^T \log P(s_t | s_{1}, \ldots, s_{t-1}).
\label{eq:loss}
\end{equation}
The recurrence is unrolled backwards in time using the back-propagation through time (BPTT) algorithm~\cite{BPTT1988}, and gradients are accumulated over multiple time-steps.

\begin{figure}
\centering
\includegraphics[scale=0.65]{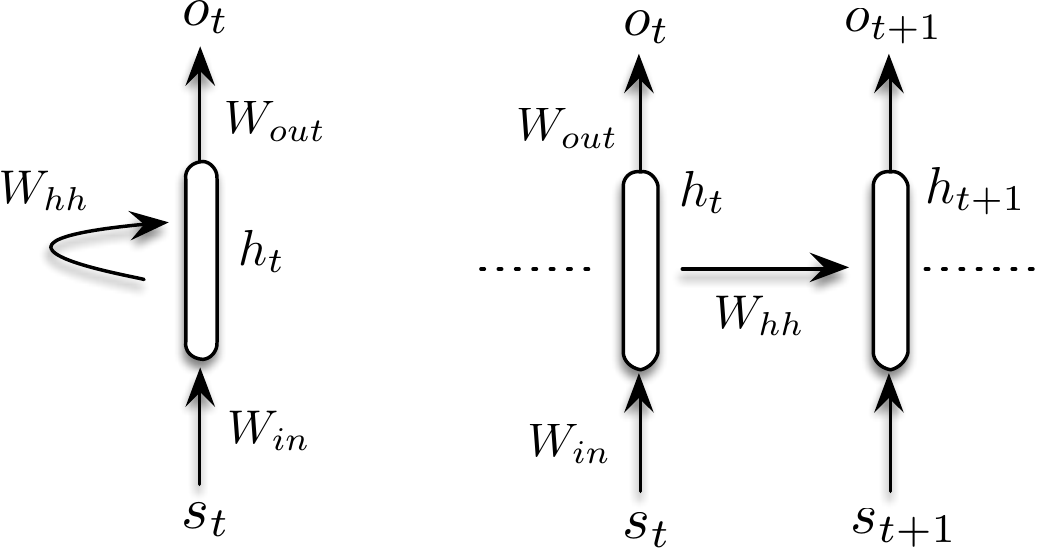}
\caption{Compact representation of an RLM (left) and unrolled representation for two time steps (right).
}
\end{figure}

\section{Context-Sensitive Models}
We distinguish three linguistic entities in a conversation between two users $A$ and $B$: the context\footnote{In this work, the context is purely linguistic, but future work might integrate further contextual information, e.g., geographical location, time information, or other forms of grounding.} $c$, the message $m$ and response $r$.
The context $c$ represents a sequence of past dialog exchanges of any length; then $B$ emits a message $m$ to which $A$ reacts by formulating its response $r$ (see Figure~\ref{tab:sample_triple}).

We use three context-based generation models to estimate a generation model of the response $r$, $r = r_1, \ldots, r_T$, conditioned on past information $c$ and $m$:
\begin{equation}
p(\B{r} | \B{c}, \B{m}) = \prod_{t = 1}^T p(r_t | r_{1}, \ldots, r_{t-1}, \B{c}, \B m).
\end{equation}
These three models differ in the manner in which they compose the context-message pair $(c, m)$.

\subsection{\label{sec:rltm}Tripled Language Model}
In our first model, dubbed RLMT, we straightforwardly concatenate each utterance $c$, $m$, $r$ into a single sentence $\B s$ and train the RLM to minimize $L(\B s)$. 
Given $c$ and $m$, we compute the probability of the response as follows: we perform the forward propagation over the known utterances $\B c$ and $m$ to obtain a hidden state encoding useful information about previous utterances. 
Subsequently, we compute the likelihood of the response from that hidden state.

An issue with this simple approach is that the concatenated sentence $s$ will be very long on average, especially if the context comprises multiple utterances.
Modelling such long-range dependencies with an RLM is difficult and is still considered an open problem~\cite{Pascanu2013}.
We will consider RLMT as an additional context-sensitive baseline for the models we present next.

\subsection{Dynamic-Context Generative Model I}
\begin{figure}
\centering
\includegraphics[scale=0.7]{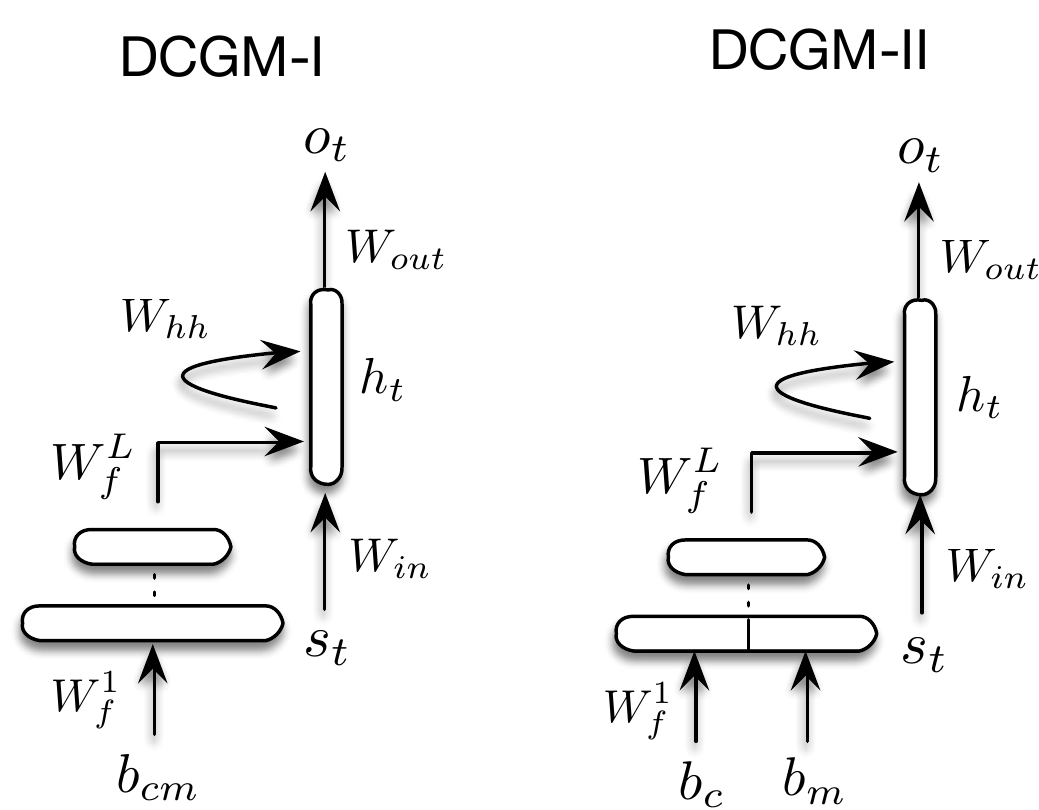}
\caption{\label{fig:dcgm}Compact representations of DCGM-I (left) and DCGM-II (right). The decoder RLM receives a bias from the context encoder. In DCGM-I, we encode the bag-of-words representation of both $\B c$ and $\B m$ in a single vector $\B b_{cm}$. In DCGM-II, we concatenate the representations $\B b_c$ and $\B b_m$ on the first layer to preserve order information.}
\end{figure}

The above limitation of RLMT can be addressed by strengthening the context bias. In our second model (DCGM-I), the context and the message are encoded into a fixed-length vector representation the is  used by the RLM to decode the response.
This is illustrated in Figure~\ref{fig:dcgm} (left). 
First, we consider $\B c$ and $\B m$ as a single sentence and compute a single bag-of-words representation $b_{cm} \in \mathbb{R}^{V}$. 
Then, $b_{cm}$ is provided as input to a multilayered non-linear forward architecture that produces a fixed-length representation that is used to bias the recurrent state of the decoder RLM.
At training time, both the context encoder and the RLM decoder are learned so as to minimize the negative log-probability of the generated response.

The parameters of the model are $\Theta_{\text{DCGM-I}} = \langle \B W_{in}, \B W_{hh}, \B W_{out}, \{ \B W_f^{\ell} \}_{\ell=1}^L \rangle$, where $\{ \B W_f^{\ell} \}_{\ell=1}^L$ are the weights for the $L$ layers of the feed-forward context networks. 
The fixed-length context vector $\B k_L$ is obtained by forward propagation of the network:
\begin{equation}
\begin{array}{c}
\B k_{1} = b_{cm}^\top W_f^1 \vspace{1mm}\\
\B k_{\ell} = \sigma(\B k_{\ell-1}^\top \B W_f^\ell) \hspace{5mm} \text{for} \; \ell = 2, \cdots, L
\end{array}
\end{equation}
The rows of $\B{W}_f^1$ contain the embeddings of the vocabulary.\footnote{Notice that the first layer of the encoder network is linear. We found that this helps learning  the embedding matrix as it reduces the vanishing gradient effect partially due to stacking of squashing non-linearities~\cite{Pascanu2013}.}
These are different from those employed in the RLM and play a crucial role in promoting the specialization of the context encoder to a distinct task. The hidden layer of the decoder RLM takes the following form:
\begin{subequations}
\begin{equation}
\B{h}_{t} = \sigma(\B{h}_{t-1}^\top \B{W}_{hh} + \B{k}_{L} + \B{s}_{t}^\top \B{W}_{in}) 
\end{equation}
\begin{equation}
\B o_t = \B h_t^\top \B W_{out}
\end{equation}
\begin{equation}
\B p(s_{t+1} | s_{1}, \ldots, s_{t-1}, c, m) = \text{softmax}(o_t)
\end{equation}
\end{subequations}
This model conditions on the previous utterances via biasing the hidden layer state on the context representation $k_L$. 
Note that the context representation does not change through time. 
This is useful because: (a) it forces the context encoder to produce a representation general enough to be useful for generating all words in the response and (b) it helps the RLM decoder to remember context information when generating long responses.

\subsection{Dynamic-Context Generative Model II}
Because DCGM-I does not distinguish between $c$ and $m$, that model has the propensity to underestimate the strong dependency that holds between $m$ and $r$.  Our third model (DCGM-II) addresses this issue by concatenating the two linear mappings of the bag-of-words representations $b_c$ and $b_m$ in the input layer of the feed-forward network representing $c$ and $m$ (see Figure~\ref{fig:dcgm} right).
Concatenating continuous representations prior to deep architectures is a common strategy to obtain order-sensitive representations~\cite{Bengio2003,Devlin2014}. 

The forward equations for the context encoder are:
\begin{equation}
\begin{array}{c}
\B k_1 = [\B b_c^\top W_f^{1}, \B b_m^\top W_f^{1}], \\ 
\B k_\ell = \sigma (\B k_{\ell-1}^\top W_f^{\ell} ) \hspace{5mm} \text{for} \; \ell = 2, \cdots, L \\
\end{array}
\end{equation}
where $[\B x, \B y]$ denotes the concatenation of $\B x$ and $\B y$ vectors. 
In DCGM-II, the bias on the recurrent hidden state and the probability distribution over the next token are computed as described in Eq. 7.
\section{Experimental Setting}

\subsection{Dataset Construction}
For computational efficiency and to alleviate the burden of human evaluators, we restrict the context sequence $c$ to a single sentence. Hence, our dataset is composed of ``triples'' $\tau \equiv (c_\tau, m_\tau, r_\tau)$ consisting of three sentences.
We mined 127M context-message-response triples from the Twitter FireHose, covering the 3-month period June 2012 through August 2012. Only those triples where context and response were generated by the same user were extracted.  To minimize noise, we selected triples that contained at least one frequent bigram that appeared more than 3 times in the corpus. This produced a corpus of 29M Twitter triples.
Additionally, we hired crowdsourced raters to evaluate approximately 33K candidate triples. 
Judgments on a 5-point scale were obtained from 3 raters apiece. 
This yielded a set of 4232 triples with a mean score of 4 or better that was then randomly binned into a tuning set of 2118 triples and a test set of 2114 triples\footnote{The Twitter ids of the tuning and test sets along with the code for the neural network models may be obtained from http://research.microsoft.com/convo/}.
The mean length of responses in these sets was approximately 11.5 tokens, after cleanup (e.g., stripping of emoticons), including punctuation.

\subsection{Automatic Evaluation}
We evaluate all systems using BLEU~\cite{BLEU} and METEOR~\cite{METEOR}, and supplement these results with more targeted
human pairwise comparisons in Section~\ref{sec:humaneval}.
A major challenge in using these automated metrics for response generation is that the set of reasonable responses in our task is potentially vast and extremely diverse. 
The dataset construction method just described yields only a single reference for each status. Accordingly, we extend the set of references using an IR approach to mine potential responses, after which we have human judges rate their appropriateness. 
As we see in Section~\ref{sec:humaneval},  it turns out that by optimizing systems towards BLEU using mined multi-references, BLEU rankings align well with human judgments. 
This lays groundwork for interesting future correlation studies.


\begin{table}[tbp]
\small
\centering
\renewcommand{\arraystretch}{1}
\begin{tabular}{@{}lccc@{}}
\toprule
\B {\bf Corpus} & {\bf \# Triples} & {\bf Avg \# Ref} & {\bf [Min,Max] \# Ref}\\
\midrule
Tuning & 2118 & 3.22 & [1, 10] \\
Test &  2114 & 3.58 & [1, 10] \\
\bottomrule
\end{tabular}
\caption{
\label{tab:multiref}
Number of triples, average, minimum and maximum number of references for tuning and test corpora.}
\end{table}

\paragraph{Multi-reference extraction} 
We use the following algorithm to better cover the space of reasonable responses. 
Given a test triple $\tau \equiv (c_\tau, m_\tau, r_\tau )$, our goal is to mine other responses $\{ r_{\tilde\tau} \}$ that fit the context and message pair $( c_\tau, m_\tau )$.
To this end, we first select a set of 15 candidate triples $\{\tilde\tau\}$ using an IR system.
The IR system is calibrated in order to select candidate triples $\tilde\tau$ for which both the message $m_{\tilde\tau}$ and the response $r_{\tilde\tau}$ are similar to the original message $m_{\tau}$ and response $r_{\tau}$.
Formally, the score of a candidate triple is:
\begin{equation}
s(\tilde\tau, \tau) = d(m_{\tilde\tau}, m_\tau)\, ( \alpha\,
d(r_{\tilde\tau}, r_\tau ) + (1 - \alpha) \epsilon),
\end{equation}
where $d$ is the bag-of-words BM25 similarity function~\cite{Robertson1995}, $\alpha$ controls the impact of the similarity between the responses and $\epsilon$ is a smoothing factor that avoids zero scores for candidate responses that do not share any words with the reference response.
We found that this simple formula provided references that were both diverse and plausible. 
Given a set of candidate triples $\{\tilde\tau\}$, human evaluators are asked to rate the quality of the response within the new triples $\{(c_\tau, m_\tau, r_{\tilde\tau})\}$.
After human evaluation, we retain the references for which the score is 4 or better on a 5 point scale, resulting in 3.58 references per example on average (Table~\ref{tab:multiref}). 
The average lengths for the responses in the multi-reference tuning and test sets are 8.75 and 8.13 tokens respectively.

\subsection{Feature Sets}
The response generation systems evaluated in this paper are parameterized as log-linear models in a framework typical of statistical machine translation~\cite{OchNey2004}.
These log-linear models comprise the following feature sets:

\label{sec:models}
\paragraph{MT}
MT features are derived from a large response generation system built along the lines of~\newcite{Ritter2011}, which is based on a phrase-based MT decoder similar to Moses~\cite{Koehn2007}. 
Our MT feature set includes the following features that are common in Moses: forward and backward maximum likelihood ``translation'' probabilities, word and phrase penalties, linear distortion, and a modified Kneser-Ney language model \cite{Kneser1995} trained on Twitter responses.
For the translation probabilities, we built a very large phrase table of 160.7 million entries by first filtering out Twitterisms (e.g., long sequences of vowels, hashtags), and then selecting candidate phrase pairs using Fisher's exact test~\cite{Ritter2011}. 
We also included MT decoder features specifically motivated by the response generation task: Jaccard distance between source and target phrase, Fisher's exact probability, and a score relating the lengths of source and target phrases.

\paragraph{IR} 
\label{par:ir}
We also use an IR feature built from an index of triples, whose implementation roughly matches the IR$_\text{status}$ approach described in~\newcite{Ritter2011}: For a test triple $\tau$, we choose $r_{\tilde\tau}$ as the candidate response iff $\tilde\tau = \arg\max_{\tilde\tau} d(m_\tau, m_{\tilde\tau})$.

\paragraph{\text{CMM}} 
Neither MT nor IR traditionally take into account contextual information. 
Therefore, we take into consideration context and message matches (CMM), i.e., exact matches between $c$, $m$ and $r$. 
We define $8$ features as the [1-4]-gram matches between $c$ and the candidate reply $r$ and the [1-4]-gram matches between $m$ and the candidate reply $r$. 
These exact matches help capture and promote contextual information in the replies.

\paragraph{RLMT, DCGM-I, DCGM-II} 
We consider the RLM trained on the concatenated triples, denoted as RLMT (Section~\ref{sec:rltm}), to be a context-sensitive RLM baseline. Each neural network model contributes an additional feature corresponding to the likelihood of the candidate response given context and message.

\begin{table}[tbp]
\centering
\renewcommand{\arraystretch}{1}
\begin{tabular}{@{}lrr@{}}
\toprule
\B {\bf System} & \hspace{2cm} & \B {\bf BLEU}\\
\midrule
{\sc random} & & 0.33 \\
{\sc mt} & & 3.21 \\
{\sc human}  &  & 6.08 \\
\bottomrule
\end{tabular}
\caption{
\label{tab:managing}
Multi-reference corpus-level BLEU obtained by leaving one reference out at random.
}
\end{table}
\begin{table*}[htp]
\begin{minipage}{.5\linewidth}
\small
\centering
\begin{tabular}{@{}lll@{}}
\toprule
{\bf MT $n$-best} & {\bf BLEU \scriptsize($\%$)} & {\bf METEOR \scriptsize($\%$)} \\
\midrule
MT$_{\text{ 9 feat.}}$ & 3.60 \scriptsize(-\emph{9.5\%}) & 9.19 \scriptsize(\emph{-0.9\%}) \\
{CMM}$_{\text{ 9 feat.}}$ & 3.33 \scriptsize(-\emph{16\%}) & 9.34 \scriptsize(\emph{+0.7\%}) \\
$\rhd$ MT + CMM$_{\text{ 17 feat.}}$ & 3.98 (-) & 9.28 (-)\\
\midrule
RLMT$_{\text{ 2 feat.}}$ &  4.13 \scriptsize(\emph{+3.7\%}) & 9.54 \scriptsize(\emph{+2.7\%}) \\
DCGM-I$_{\text{ 2 feat.}}$  & 4.26 \scriptsize(\emph{+7.0\%}) & 9.55 \scriptsize(\emph{+2.9\%}) \\
DCGM-II$_{\text{ 2 feat.}}$  &  4.11 \scriptsize(\emph{+3.3\%}) & 9.45 \scriptsize(\emph{+1.8\%}) \\
\midrule
DCGM-I + CMM$_{\text{ 10 feat.}}$ &  4.44 \scriptsize(\emph{+11\%}) & 9.60 \scriptsize(\emph{+3.5\%}) \\
DCGM-II + CMM$_{\text{ 10 feat.}}$&  4.38 \scriptsize(\emph{+10\%}) & 9.62 \scriptsize(\emph{+3.5\%}) \\
\bottomrule
\end{tabular}
\end{minipage}%
\begin{minipage}{.5\linewidth}
\small
\centering
\begin{tabular}{@{}lll@{}}
\toprule
{\bf IR $n$-best} & {\bf BLEU \scriptsize($\%$)} & {\bf METEOR \scriptsize($\%$)} \\
\midrule
IR$_{\text{ 2 feat.}}$  & 1.51 \scriptsize(\emph{-55\%}) & 6.25 \scriptsize(\emph{-22\%}) \\
CMM$_{\text{ 9 feat.}}$ & 3.39 \scriptsize(\emph{-0.6\%}) & 8.20 \scriptsize(\emph{+0.6\%}) \\
$\rhd$  IR + CMM$_{\text{ 10 feat.}}$ & 3.41 (-) & 8.04 (-) \\
\midrule
RLMT$_{\text{ 2 feat.}}$ & 2.85 \scriptsize(\emph{-16\%}) & 7.38 \scriptsize(\emph{-8.2\%}) \\
DCGM-I$_{\text{ 2 feat.}}$ & 3.36 \scriptsize(\emph{-1.5\%}) & 7.84 \scriptsize(\emph{-2.5\%})\\
DCGM-II$_{\text{ 2 feat.}}$ & 3.37 \scriptsize(\emph{-1.1\%})& 8.22 \scriptsize(\emph{+2.3\%})\\
\midrule
DCGM-I + CMM$_{\text{ 10 feat.}}$  & 4.07 \scriptsize(\emph{+19\%})& 8.67 \scriptsize(\emph{+7.8\%})\\
DCGM-II + CMM$_{\text{ 10 feat.}}$  & 4.24 \scriptsize(\emph{+24\%})& 8.61 \scriptsize(\emph{+7.1\%})\\
\bottomrule
\end{tabular}
\end{minipage}
\caption{
\label{tab:ranking}
Context-sensitive ranking results on both MT (left) and IR (right) $n$-best lists, $n = 1000$. 
The subscript $_\text{feat.}$ indicates the number of features of the models. The log-linear weights are estimated by running one iteration of MERT. We mark by ($\pm$\%) the relative improvements with respect to the reference system ($\rhd$).
}
\end{table*}
\subsection{Model Training}
The proposed models are trained on a 4M subset of the triple data. The vocabulary consists of the most frequent $V = 50K$ words. 
In order to speed up training, we use the Noise-Contrastive Estimation (NCE) loss, which avoids repeated summations over $V$ by approximating the probability of the target word~\cite{Gutmann2010}.
Parameter optimization is done using Adagrad~\cite{Duchi2011} with a mini-batch size of 100 and a learning rate $\alpha = 0.1$, which we found to work well on held-out data. 
In order to stabilize learning, we clip the gradients to a fixed range $[-10, 10]$, as suggested in~\newcite{Mikolov2010}. All the parameters of the neural models are sampled from a normal distribution $\mathcal{N}(0, 0.01)$ while the recurrent weight $W_{hh}$ is initialized as a random orthogonal matrix and scaled by 0.01. To prevent over-fitting, we evaluate performance on a held-out set during training and stop when the objective increases. 
The size of the RLM hidden layer is set to $K = 512$, where the context encoder is a $512$, $256$, $512$ multilayer network.
The bottleneck in the middle compresses context information that leads to similar responses and thus achieves better generalization. The last layer embeds the context vector into the hidden space of the decoder RLM.

\subsection{Rescoring Setup}
\label{sec:rescoring}
We evaluate the proposed models by rescoring the $n$-best candidate responses obtained using the MT phrase-based decoder and the IR system.
In contrast to MT, the candidate responses provided by IR have been created by humans and are less affected by fluency issues.
The different $n$-best lists will provide a comprehensive testbed for our experiments.
First, we augment the $n$-best list of the tuning set with the scores of the model of interest.
Then, we run an iteration of MERT~\cite{Och2003} to estimate the log-linear weights of the new features.
At test time, we rescore the test $n$-best list with the new weights.

\section{Results}
\subsection{Lower and Upper Bounds}
Table 2 shows the expected upper and lower bounds for this task as suggested by BLEU scores for human responses and a random response baseline. The {\sc random} system comprises responses randomly extracted from the triples corpus.
{\sc human} is computed by choosing one reference amongst the
multi-reference set for each context-status pair.\footnote{For the human score, we compute corpus-level BLEU with a sampling scheme that randomly leaves out one reference - the human sentence to score - for each reference set. This sampling scheme (repeated with 100 trials) is also applied for the MT and {\sc random} system so as to make BLEU scores comparable.}  
Although the scores are lower than those usually reported in SMT tasks, the ranking of the three systems is unambiguous.

\subsection{BLEU and METEOR}
The results of automatic evaluation using BLEU and METEOR are presented in Table~\ref{tab:ranking}, where some broad patterns emerge.  First,  both metrics indicate that a phrase-based MT decoder outperforms a purely IR approach.  Second, adding CMM features to the baseline systems helps.  Third, the neural network models contribute measurably to improvement:  RLMT and DCGM models outperform baselines, and DCGM models provide more consistent gains than RLMT.

\paragraph{MT vs.\ IR}
\label{par:mtvsir}
BLEU and METEOR scores indicate that the phrase-based MT decoder outperforms a purely IR approach, despite the fact that IR proposes fluent human generated responses.  This may be because the IR model only loosely captures important patterns between message and response: It ranks candidate responses solely by the similarity of their message with the message of the test triple (\S\ref{sec:models}). As a result, the top ranked response is likely to drift from the purpose of the original conversation. The MT approach, by contrast, more directly models statistical patterns between message and response.

\begin{figure}
\centering
\includegraphics[scale=0.40]{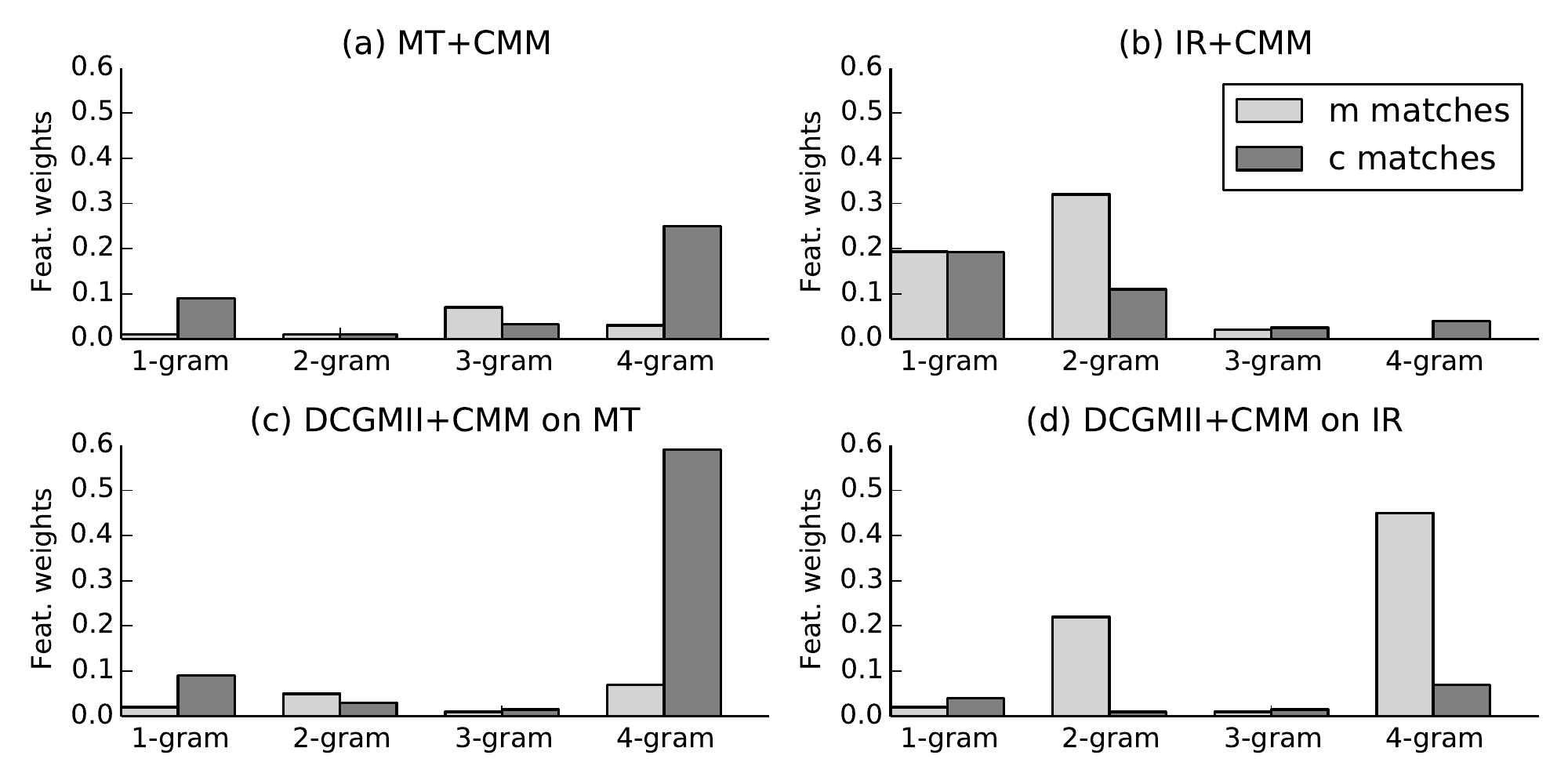}
\caption{
\label{fig:mtir}
Comparison of the weights of learned CMM features for MT+CMM and IR+CMM systems (a) et (b) and DCGM-II+CMM on MT and IR (c) and (d).
}
\end{figure}

\paragraph{CMM}
\label{par:cmmfeat}
MT+CMM, totaling 17 features (9 from MT + 8 CMM), improves 0.38 BLEU points, a $9.5\%$ relative improvement, over the baseline MT model.  
IR+CMM, with 10 features (IR + word penalty + 8 CMM), benefits even more, attaining 1.8 BLEU points and 1.5 METEOR points over the IR baseline.  
Figure~\ref{fig:mtir} (a) and (b) plots the magnitude of the learned CMM feature weights for MT+CMM and IR+CMM.  
CMM features help in both these hypothesis spaces and especially on the IR $n$-best list.
Figure~\ref{fig:mtir} (b) supports the hypothesis formulated in the previous paragraph: Since IR solely captures inter-message similarities, the matches between message and response are important, while context matches help in providing additional gains.
The phrase-based statistical patterns captured by the MT system do a good job in explaining away 1-gram and 2-gram message matches (Figure~\ref{fig:mtir} (a)) and the performance gain mainly comes from context matches. On the other hand, we observe that 4-gram matches may be important in selecting appropriate responses. Inspection of the tuning set reveals instances where responses contain long subsequences of their corresponding messages, e.g., $m$ = ``good night best friend, I love you'', $r$ = ``I love you too, good night best friend''.  Although infrequent, such higher-order n-gram matches, when they occur, may provide a more robust signal of the quality of the response than 1- and 2-gram matches, given the highly conversational nature of our dataset.

\paragraph{RLMT and DCGM}
\label{par:rlmtdcgm}
Both RLMT and DCGM models outperform their respective MT and IR baselines.
Both models also exhibit similar performance and show improvements over the MT+CMM models, albeit using a lower dimensional feature space.
We believe that their similar performance is due to the limited diversity of MT $n$-best list together with gains in fluency stemming from the strong language model provided by the RLM.
In the case of IR models, on the other hand, there is more headroom for improvement and fluency is already guaranteed. 
Any gains must come from context and message matches. 
Hence, RLMT underperforms with respect to both DCGM and IR+CMM.
The DCGM models appear to have better capacity to retain contextual information and thus achieve similar performance to IR+CMM despite their lack of exact n-gram match information.

In the present experimental setting, no striking performance difference can be observed between the two versions of the DCGM architecture. If multiple sequences were used as context, we expect that the DCGM-II model would likely benefit more owing to the separate encoding of message and context.

\begin{table}[tbp]
\small
\centering
\begin{tabular}{@{}llcc@{}}
\toprule
{\bf System A} & {\bf System B} & {\bf Gain (\%)} & {\bf CI}\\
\midrule
HUMAN & MT+CMM & 13.6* & [12.4,14.8] \\
\midrule
DCGM-II & MT & 1.9* & [0.8, 2.9] \\
DCGM-II+CMM & MT & 3.1* & [2.0, 4.3] \\
DCGM-II+CMM & MT+CMM & 1.5* & [0.5, 2.5] \\
\midrule
DCGM-II & IR & 5.2* & [4.0, 6.4] \\
DCGM-II+CMM & IR & 5.3* & [4.1, 6.6] \\
DCGM-II+CMM & IR+CMM & 2.3* & [1.2, 3.4] \\
\bottomrule
\end{tabular}
\caption{\label{tab:humanscores} Pairwise human evaluation scores between System A and B. The first (second) set of results refer to the MT  (IR) hypothesis list. The asterisk means agreement between human preference and BLEU rankings.}
\end{table}

\paragraph{DCGM+CMM}
\label{par:dcgmcmm}
We also investigated whether mixing exact CMM n-gram overlap with semantic information encoded by the DCGM models can bring additional gains. 
DCGM-\{I-II\}+CMM systems each totaling 10 features show increases of up to 0.48 BLEU points over MT+CMM and up to 0.88 BLEU over the model based on~\newcite{Ritter2011}.
METEOR improvements similarly align with BLEU improvements both for MT and IR lists.
We take this as evidence that CMM exact matches and DCGM semantic matches interact positively, a finding that comports with~\newcite{Gao2014}, who show that semantic relationships mined through phrase embeddings correlate positively with classic co-occurrence-based estimations.
Analysis of CMM feature weights in Figure~\ref{fig:mtir} (c) and (d) suggests that 1-gram matches are explained away by the DCGM model, but that higher order matches are important.
It appears that DCGM models might be improved by preserving word-order information in context and message encodings.

\subsection{Human Evaluation}
\label{sec:humaneval}
Human evaluation was conducted using crowdsourced annotators. Annotators were asked to compare the quality of system output responses pairwise (``Which is better?") in relation to the context and message strings in the 2114 item test set.  Identical strings were held out, so that the annotators only saw those outputs that differed. Paired responses were presented in random order to the annotators, and each pair of responses was judged by 5 annotators.

Table 4 summarizes the results of human evaluation, giving the difference in mean scores (pairwise preference margin) between systems and 95\% confidence intervals generated using Welch's t-test. Identical strings not shown to raters are incorporated with an automatically assigned score of 0.5. The pattern in these results is clear and consistent: context-sensitive systems (+CMM) outperform non-context-sensitive systems, with preference gains as high as approximately 5.3\% in the case of DCGM-II+CMM versus IR, and about 3.1\% in the case of DCGM-II+CMM versus MT. 
Similarly, context-sensitive DCGM systems outperform non-DCGM context-sensitive systems by 1.5\% (MT) and 2.3\% (IR). 
These results are consistent with the automated BLEU rankings and confirm that our best performing DCGM models outperform both raw baseline and the context-sensitive baseline using CMM features. 
\subsection{Discussion}

\begin{table*}[tpb]
\centering
\scriptsize
\renewcommand{\arraystretch}{1.1}
\begin{tabulary}{\linewidth}{@{}c@{\hspace{1em}}L@{\hspace{1em}}L@{\hspace{1em}}L@{}}
\toprule
{\bf \#} &  {\bf Context $c$} & {\bf Message $m$} & {\bf Generated Response $r$}\\
\midrule
1 & noone can help this time its just i need a support system & well im here to support you. whenever you need me. & thank you, i appreciate it thank you \\
\midrule
2 & nothin much, and how's the book?! & its good but i'm only like halfway through cuz i don't feel like reading. i'm so bored ... & that's good! i have the book but i'm bored too. \\
\midrule
3 & i know! what have you been up to? & nothing really. just enjoying summer and working. what about you? & that's good! nothing much just listening to music and you? \\
\midrule
4 & thanks, your way better though & i'm just good at expressing it that's but trust me you are freakin amazing and waay better & i know, i don't trust you, i love you too! thank you \\
\midrule
5 & thanks. yeah i am good thanks! good, but angry! & sorry you're angry, been there, i hope it dies down. & i hope so too, i'm not angry!! \\
\midrule
6 & figures. bass. never ate bass. & bass is extremely mild with very few bones. my favorite after catfish & that's one of the bones, chuck bass is mine too.\\
\midrule
7 & today i will walk for miles. clear sky clear mind. & love walking. it's too hot to walk this month. i haven't moved. frozen. heat freeze. & it's too hot for you to get up to honey if you have frozen yogurt to you.\\
\midrule
8 & arresting people over ``nasty tweets"? please. i'm not condoning what the lad did but arresting him is a tad silly. & it appears he may have sent death threats to others while all this was kicking off  too. & i did! he was armed with nuclear threats?  that's what's happening to you.\\
\bottomrule
\end{tabulary}
\caption{
\label{tab:examples}
Sample responses produced by the MT-based DCGM-II+CMM system. }
\end{table*}

Table~\ref{tab:examples} provides examples of responses generated on the tuning corpus by the MT-based DCGM-II+CMM system, our best system in terms of both BLEU and human evaluation. Responses from this system are on average shorter (8.95 tokens) than the original human responses in the tuning set (11.5 tokens). Overall, the outputs tend to be generic or commonplace, but are often reasonably plausible in the context as in examples 1-3, especially where context and message contain common conversational elements.  Example 2 illustrates the impact of context-sensitivity: the word ``book'' in the response is not found in the message. Nonetheless, longer generated responses are apt to degrade both syntactically and in terms of content. We notice that longer responses are likely to present information that conflicts either internally within the response itself, or is at odds with the context, as in examples 4-5. This is not unsurprising, since our model lacks mechanisms both for reflecting agent intent in the response and for maintaining consistency with respect to sentiment polarity. Longer context and message components may also result in responses that wander off-topic or lapse into incoherence as in 6-8, especially when relatively low frequency unigrams (``bass'', ``threat'') are echoed in the response. In general, we expect that larger datasets and incorporation of more extensive contexts into the model will help yield more coherent results in these cases. Consistent representation of agent intent is outside the scope of this work, but will likely remain a significant challenge.


\section{Conclusion}

We have formulated a neural network architecture for data-driven response generation trained from social media conversations, in which generation of responses is conditioned on past dialog utterances that provide contextual information. 
We have proposed a novel multi-reference extraction technique allowing for robust automated evaluation using standard SMT metrics such as BLEU and METEOR.
Our context-sensitive models consistently outperform both context-independent and context-sensitive baselines by up to 11\% relative improvement in BLEU in the MT setting and 24\% in the IR setting, albeit using a minimal number of features.
As our models are completely data-driven and self-contained, they hold the potential to improve fluency and contextual relevance in other types of dialog systems.

Our work suggests several directions for future research. 
We anticipate that there is much room for improvement if we employ more complex neural network models that take into account word order within the message and context utterances. 
Direct generation from neural network models is an interesting and potentially promising next step.
Future progress in this area will also greatly benefit from thorough study of automated evaluation metrics.

%

\section*{Acknowledgments}
We thank Alan Ritter, Ray Mooney, Chris Quirk, Lucy Vanderwende, Susan Hendrich and Mouni Reddy for helpful discussions, as well as the three anonymous reviewers for their comments.

\bibliographystyle{naaclhlt2015}
\bibliography{naacl15-sordoni}
\end{document}